\documentclass[journal,twocolumn]{IEEEtran}
\usepackage{cite}
\usepackage{amsmath,amssymb,amsfonts}
\usepackage{algorithmic}
\usepackage{graphicx}
\usepackage{textcomp}
\usepackage{booktabs}
\usepackage{comment}
\usepackage{pifont}
\usepackage{threeparttable}

\def\BibTeX{{\rm B\kern-.05em{\sc i\kern-.025em b}\kern-.08em
    T\kern-.1667em\lower.7ex\hbox{E}\kern-.125emX}}
\begin{document}
\title{Real-Time Pulsatile Flow Prediction for Realistic, Diverse Intracranial Aneurysm Morphologies using a Graph Transformer and Steady-Flow Data Augmentation}
\author{Yiying Sheng\textsuperscript{*}, Wenhao Ding\textsuperscript{*}, Dylan Roi, Leonard Leong Litt Yeo, Hwa Liang Leo, and Choon Hwai Yap
\thanks{\textsuperscript{*}These authors contributed equally to this work.}
\thanks{Y. Sheng is with the Department of Biomedical Engineering, National University of Singapore, 4 Engineering Drive 3, Singapore 117583 (e-mail: yiyings@nus.edu.sg).}
\thanks{W. Ding is with the Department of Biomedical Engineering, Imperial College London, White City Campus, London W12 0BZ, U.K. (e-mail: w.ding23@imperial.ac.uk).}
\thanks{D. Roi is with Imperial College Healthcare NHS Trust, London W6 8RF, U.K. (e-mail: dylan.roi@nhs.net).}
\thanks{L. LL. Yeo is with the Department of Medicine, National University Hospital, 1E Kent Ridge Road, Singapore 119228. (e-mail: leonard\_ll\_yeo@nuhs.edu.sg).}
\thanks{H. L. Leo is with the Department of Biomedical Engineering, National University of Singapore, 4 Engineering Drive 3, Singapore 117583 (e-mail: bielhl@nus.edu.sg).}
\thanks{C. H. Yap is with the Department of Biomedical Engineering, Imperial College London, White City Campus, London W12 0BZ, U.K. (e-mail: c.yap@imperial.ac.uk).}}

\maketitle

\begin{abstract}
Extensive studies suggested that fluid mechanical markers of intracranial aneurysms (IAs) derived from Computational Fluid Dynamics (CFD) can indicate disease progression risks, but to date this has not been translated clinically. This is because CFD requires specialized expertise and is time-consuming and low throughput, making it difficult to support clinical trials. A deep learning model that maps IA morphology to biomechanical markers can address this, enabling physicians to obtain these markers in real time without performing CFD. Here, we show that a Graph Transformer model that incorporates temporal information, which is supervised by large CFD data, can accurately predict Wall Shear Stress (WSS) across the cardiac cycle from IA surface meshes. The model effectively captures the temporal variations of the WSS pattern, achieving a Structural Similarity Index (SSIM) of up to 0.981 and a maximum-based relative L2 error of 2.8\%. Ablation studies and SOTA comparison confirmed its optimality. Further, as pulsatile CFD data is computationally expensive to generate and sample sizes are limited, we engaged a strategy of injecting a large amount of steady-state CFD data, which are extremely low-cost to generate, as augmentation. This approach enhances network performance substantially when pulsatile CFD data sample size is small. Our study provides a proof of concept that temporal sequences cardiovascular fluid mechanical parameters can be computed in real time using a deep learning model from the geometric mesh, and this is achievable even with small pulsatile CFD sample size. Our approach is likely applicable to other cardiovascular scenarios.
\end{abstract}


\section{Introduction}
\label{sec:introduction}
Intracranial aneurysms (IAs) are abnormal dilations of intracranial arteries, affecting 2-5\% of global population \cite{rayz2020}. Risk factors include familial predisposition, hypertension, smoking, and increasing age \cite{brown2014}. While most aneurysms are asymptomatic, their potential for rupture represents a significant threat to public health. A ruptured IA can cause subarachnoid hemorrhage (SAH) and hemorrhagic stroke, leading to a mortality rate of 40-50\% \cite{brown2014}. Among survivors, nearly 60\% experience permanent neurological deficits and long-term disability \cite{brown2014}. Therefore, early detection and appropriate management of unruptured intracranial aneurysms (UIAs) are important.

In clinical practice, aneurysms are often discovered incidentally, except when they impinge on cranial nerves, leading to symptoms such as headaches, visual disturbance, or hemorrhage \cite{rayz2020}. Diagnosis relies on cerebrovascular imaging modalities, including digital subtraction angiography (DSA), computed tomography angiography (CTA), and magnetic resonance angiography (MRA). These techniques allow detailed assessment of aneurysm morphology (size, shape, location), which is used for rupture risk stratification to guide treatment decisions. Physicians must balance the risk of rupture to those associated with preventive interventions. While rupture can be catastrophic, surgical clipping and endovascular coiling treatment may cause severe complications, including thromboembolic events, intraoperative rupture, and delayed ischemia. 

Currently, risk stratification of UIAs mainly relies on scoring systems such as PHASES and ELAPSS. These tools use basic morphological and clinical factors, including aneurysm size, location, patient age, sex, and family history. However, these scores are often insufficient for accurate rupture risk assessment. Previous studies indicate that the PHASES score has high false-positive rate (56\%) \cite{mocco2018}, and may lead to overtreatment of stable aneurysms \cite{bijlenga2017}. Other studies report that these scoring systems fail to identify a significant proportion of ruptures in small or 'low-risk' aneurysms (false negatives) \cite{pagiola2020}.

There is thus great interest in finding new biomarkers of risk, and much attention has been given to blood flow fluid dynamics parameters, especially on fluid wall shear stresses and its derivatives. Previous in-vitro and in-vivo experiments on patient-specific models demonstrate that local hemodynamic forces can lead to wall degeneration and contribute to aneurysm instability \cite{yagi2013,benemerito2025,cebral2019}. Systematic reviews of patient-specific CFD studies have also suggested that biomechanical parameters, such as wall shear stress (WSS) and oscillatory shear index (OSI), are potential markers of instability \cite{zhou2017a,han2021}. Despite the evidence from previous studies, intra-aneurysmal flow analysis has not been adopted in routine clinical use. 4D Flow MRI can capture 3D velocity fields in vivo, but it is limited by its low spatiotemporal resolution, image artifacts and noise \cite{rayz2020}. Thus, detailed quantification of IA fluid mechanics parameters relies on CFD based on geometries extracted from vascular imaging data, which is an established methodology and is routinely performed in research studies \cite{rayz2020}. 

However, to date, CFD has not been adopted clinically to assist risk stratification. This is due to CFD being time-consuming and computationally expensive, making it prohibitive to conduct clinical trials to demonstrate the clinical utility of fluid mechanics markers, and justifying their use clinically. Another reason is that CFD requires specialized skills, which act as a strong adoption barrier by clinicians. 

To address this problem, AI-based surrogate models have been proposed in recent years to approximate CFD solutions for rapid prediction of vascular fluid mechanics. Various deep learning (DL) architectures, including convolution neural networks (CNNs) \cite{su2020a,ferdian2022,faisal2025c,gharleghi2022}, graph neural networks (GNNs) and Transformers \cite{sukMesh2022a,sukLaBGATr2024,sukGReAT2025,rygiel2025}, have been applied to learn the mapping between vascular morphology and fluid velocity, pressure, and WSS fields. However, many existing studies rely on idealized geometries or simplified 2D representations of the vasculature, or are limited to steady-state hemodynamics rather than transient flows, which is important for calculating time-derived metrics such as OSI and relative residence time (RRT). A few recent studies using graph-based networks has demonstrate the feasibility of pulsatile flow WSS prediction, but there is room for improvement to its accuracy, and there is a need for further strategies to reduce the need for computationally expensive pulsatile CFD data for supervision \cite{rygiel2025}.

In this study, we present a DL framework to predict transient WSS vector fields over a full cardiac cycle from aneurysm surface meshes, preserving the mesh structures and topological details. We train our model on the AneuG-Flow dataset \cite{ding2025aneugflow}, a large CFD simulation dataset including 14,000 steady-state cases and 808 pulsatile cases. We utilized a graph isomorphism network with edge features as the backbone, and insert both local and global attention mechanisms to enhance accuracy. Further, we adopted a strategy of using a large amount of computationally cheap steady-flow data to augment the network to enable higher accuracy with lower pulsatile CFD training data. Our ablation and SOTA comparison demonstrates the optimality of this approach. The main contributions of this work are summarized as follows: 

\begin{itemize} 
 \item We developed an DL framework for rapid computation of transient WSS from IA mesh inputs, with a GPS graph-transformer architecture utilizing a novel Graph Harmonic Deformation (GHD) shape encoding for geometric and local positioning encoding for both global and local attention. The approach more effectively interprets the mesh geometric information and avoids lengthy pre-computations during inference.
 \item We demonstrate that augmenting the graph-based pulsatile fluid mechanics prediction network with a large-scale computationally low cost steady-flow dataset can significantly improves performance for transient flow prediction with lower pulsatile sample size, thus providing a new strategy for reducing computational cost of obtaining CFD data for training. 
 \item Our use of the GHD shape encoding enables a standardized pipeline from data synthesis to network training and inference, providing computational efficiency advantages. We previously used GHD in a network for generating a realistic synthetic dataset \cite{dingGeom2026} at low computational cost to generate a large CFD dataset \cite{ding2025aneugflow}, GHD is currently used as quick but accurate down sampling of CFD mesh during training, and it is further used as inputs to our GPS network during inference as effective shape encoders that allows costly pre-computations to be avoided.
\end{itemize}

\section{Related Work}
\subsection{Deep Learning-based WSS Surrogation}
In biomedical domain, early approaches for deep learning WSS assessment mainly relied on parameterized representations of 3D geometries to apply standard CNNs. Su et al. \cite{su2020a} projected idealized stenosed coronary arteries into 2D representations to predict steady WSS. Ferdian et al. \cite{ferdian2022} unwrapped the aorta into 2D flat maps to estimate WSS from 4D Flow MRI data. However, such approaches have difficulty with representing more complex patient specific geometries such as those in IA. More recently, Faisal et al. \cite{faisal2025c} converted 3D AAA geometries to multiple 2D rendered projection images to predict time-averaged wall shear stress (TAWSS). While these projection-based methods are computationally efficient, there is inconvenience in requiring the conversion between mesh data and images, which also incurs data loss.

To overcome the limitations of 2D mappings, recent studies have adopted geometric neural architectures that operate directly on unstructured grids or point clouds. Suk et al. \cite{sukMesh2022a} introduced a mesh CNN to predict steady WSS on coronary arteries, effectively processing 3D surface mesh without requiring a parameterized 2D projection. Following this, they proposed LaB-GATr \cite{sukLaBGATr2024}, a geometric algebra transformer that tokenizes point-cloud patches on surface and volume meshes and processes them with self-attention. LaB-GATr \cite{sukLaBGATr2024} achieved state-of-the-art performance in predicting steady WSS distributions on coronary arteries and showed good generalization across different anatomies. To address data scarcity, the same research group recently introduced GReAT \cite{sukGReAT2025}, which uses self-supervised pre-training on more than 8,000 vascular shapes to improve WSS assessment in a small clinical cohort. However, this approach is limited to classifying discretized WSS categories (low/mid/high) based on time-averaged data, rather than resolving continuous WSS vector fields. Overall, despite providing improved geometric embeddings of 3D vascular anatomy, these models still focus on steady or time-averaged metrics and cannot model the transient WSS needed to derive indices like OSI and RRT.

Predicting full cardiac cycle transient WSS fields remains a significant challenge due to the high dimensionality of spatiotemporal data. Gharleghi et al. \cite{gharleghi2022} proposed a CNN framework to predict transient WSS in idealized coronary bifurcations, but their approach relies on a pre-computed steady-state WSS as input and maps the vessel geometries to 2D grids, which limits generalization to more complex coronary anatomies. Most recently, Rygiel et al. \cite{rygiel2025} used LaB-GATr to estimate transient WSS in abdominal aortic anuerysms with varying inlet flow rates. To reduce the memory cost of the transformer module, their model operates on a subsampled point cloud. While the approach is effective, it uses point-based instead of graph-based encoding and there may be room for improvement for better geometry encoding.

Specific to IAs, Li et al. \cite{liRapid2025} proposed a PointNet-based strategy for rapid WSS prediction in intracranial aneurysms; however, their model was trained on idealized aneurysm shapes, which may not capture the full complexity of patient-specific anatomies. These limitations motivate our framework for predicting full cardiac-cycle WSS fields on a large-scale, synthesized intracranial aneurysm dataset.

\begin{table}[t]
  \caption{Summary of DL-based WSS Surrogation.}
  \label{tab:literature_review}
  \centering
  \resizebox{\columnwidth}{!}{%
  \begin{tabular}{l l l c l}
    \toprule
    Study & Domain & Geometry & $N$ & Input $\rightarrow$ Output \\
    \midrule
    Su et al.~\cite{su2020a} & Coronary & Idealized & 2k & 2D Proj. $\rightarrow$ TAWSS \\
    Ferdian et al.~\cite{ferdian2022} & Aorta & Patient & 43 & 2D Map $\rightarrow$ Steady \\
    Faisal et al.~\cite{faisal2025c} & AAA & Synthetic & 253 & 2D Proj. $\rightarrow$ TAWSS \\
    Suk et al.~\cite{sukMesh2022a} & Coronary & Idealized & 2k & Mesh $\rightarrow$ Steady \\
    Suk et al.~\cite{sukLaBGATr2024} & Coronary & Idealized & 2k & Mesh $\rightarrow$ Steady \\
    Suk et al.~\cite{sukGReAT2025} & Coronary & Synthetic & $\sim$8.5k & Point Cloud $\rightarrow$ Class. \\
    Gharleghi et al.~\cite{gharleghi2022} & Coronary & Synthetic & $\sim$2.7k & 2D Grid $\rightarrow$ Transient \\
    Rygiel et al.~\cite{rygiel2025} & AAA & Patient & 100 & Features $\rightarrow$ Transient \\
    Li et al.~\cite{liRapid2025} & IA & Idealized & 2k & Point Cloud $\rightarrow$ Steady \\
    \textbf{Ours} & \textbf{IA} & \textbf{Synthetic} & \textbf{14k+808} & \textbf{Mesh $\rightarrow$ Transient} \\
    \bottomrule
  \end{tabular}%
  }
\end{table}

\subsection{Geometric Graph Learning on Meshes}
We aim to learn a neural solution operator
\begin{equation}
\mathcal{G}_\theta : (\Omega, t) \mapsto w,
\end{equation}
that maps vascular surface geometry $\Omega$ to hemodynamic fields $w(\mathbf{x}, t)$. Unlike image-based problems defined on regular Euclidean grids, vascular topologies are defined on irregular manifolds, and model performance depends on how the geometry is discretized and encoded. In practice, $\Omega$ is represented by a surface mesh $\mathcal{M}(\mathcal{V},\mathcal{F})$, and predicting WSS becomes a node-wise regression task on a non-Euclidean domain. A key challenge is to design a geometric encoder that captures both localized WSS variations and long-range interactions between the aneurysm sac and parent vessels. Existing mesh-based architectures can be broadly grouped into message passing networks \cite{pfaffMeshGraphNet2021,xuGIN2019,huGINE2020}, hierarchical pooling models \cite{gaoGraphUNets2019}, and Transformer-based models \cite{dwivediGraphTransformer2021,rampasekGraphGPS2023}.

MeshGraphNet \cite{pfaffMeshGraphNet2021} propagates information along mesh edges and preserves mesh connectivity for simulation data. However, this message-passing structure does not include a pooling mechanism and may have limited global geometric awareness. Graph Isomorphism Networks (GIN) \cite{xuGIN2019} use a sum aggregation message passing backbone to provide a strong local operator on irregular graphs. However, GIN is strictly node-centric and aggregates neighbor features without incorporating edge attributes. GIN with edge features (GINE) \cite{huGINE2020} extends GIN by conditioning messages on edges features to improve local geometry representation. However, GIN and GINE remain local operators, and capturing global context requires deep stacking of layers, which is computationally inefficient and can lead to over-smoothing on high-resolution meshes. Graph U-Net \cite{gaoGraphUNets2019} introduces pooling and unpooling to build multi-scale representations to improve global receptive field. However, pooling on unstructured grids can disrupt fine-scale correspondences and may introduce topological inconsistencies during unpooling. To capture long-range dependencies, such as the effect of the inlet flow on distal aneurysm sacs, Graph Transformers \cite{dwivediGraphTransformer2021} introduce global attention mechanisms. However, pure global attention can be expensive on dense meshes. GraphGPS \cite{rampasekGraphGPS2023} addresses this by combining local message passing with global mixing in a hybrid architecture. This motivates our use of a GPS-style backbone to capture both detailed local physics and global effects.

\section{Method}
\subsection{Overview}
We train an end-to-end model to predict the time serie of Wall Shear Stress from the aneurysm morphology. Fig.~\ref{network_schematic} demonstrates our network design, including a geometry encoding module (Section \ref{geometry_encoding}) and a time \& waveform encoding module (Section \ref{time_waveform_encoding}). Features extracted from the aneurysm morphology and the mass flow waveform signal are then fused in the feature fusion module. We train our model on the AneuG-Flow dataset \cite{ding2025aneugflow} containing 14,000 steady flow CFD simulations of varied aneurysm geometries and 808 pulsatile flow simulations on varied geometries. We test a family of baselines in our ablation experiments, and the augmentation capacity of cheap steady flow data.

\begin{figure}[!t]
\centerline{\includegraphics[width=\columnwidth]{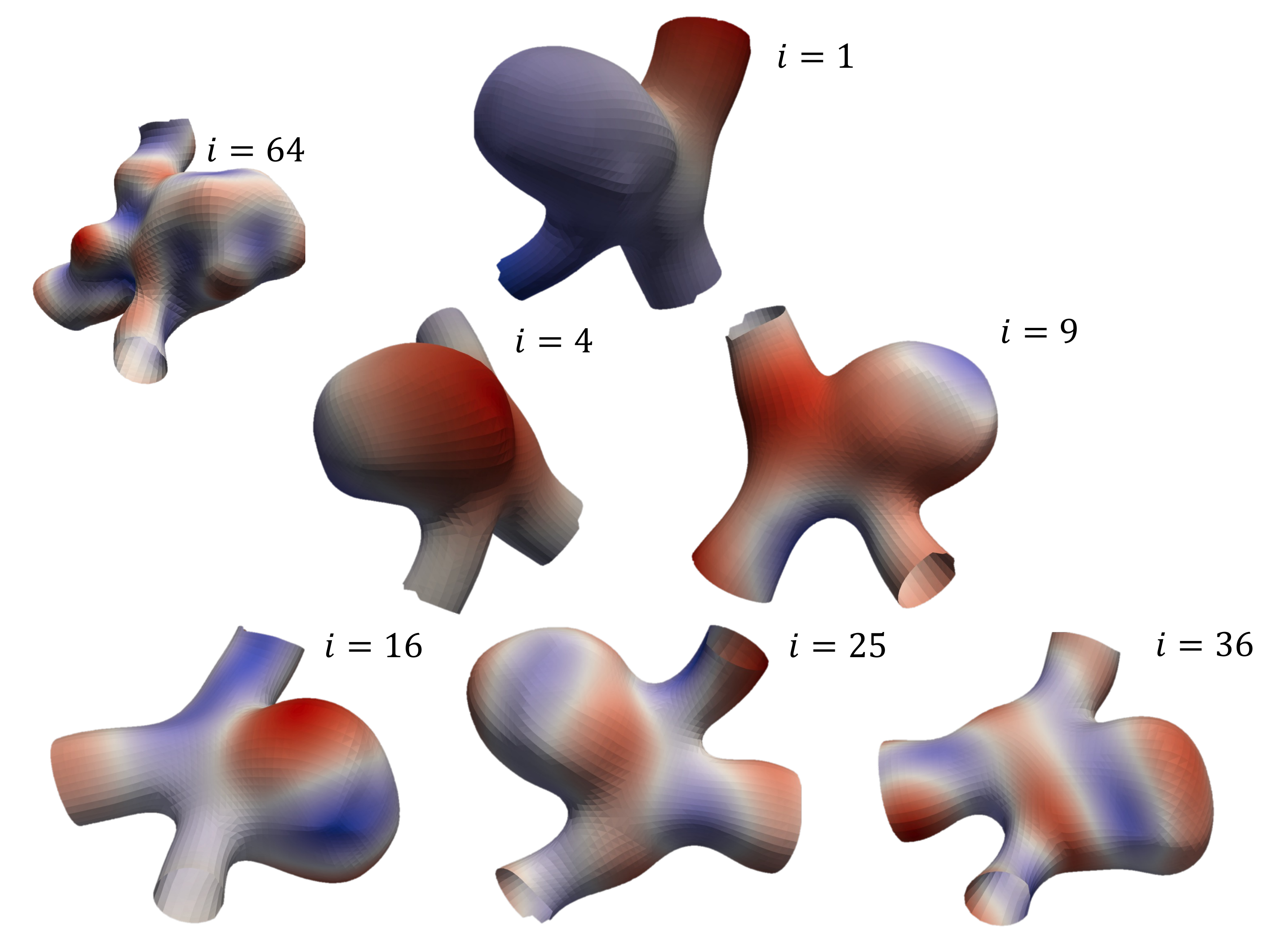}}
\caption{Graph Harmonic Deformation (GHD) shape modes. Higher modes correspond to high-frequency behaviors on the graph.}
\label{fig2}
\end{figure}

\subsection{Shape Preprocessing}
\label{shape_preprocessing}
\begin{figure}[!t]
\centerline{\includegraphics[width=\columnwidth]{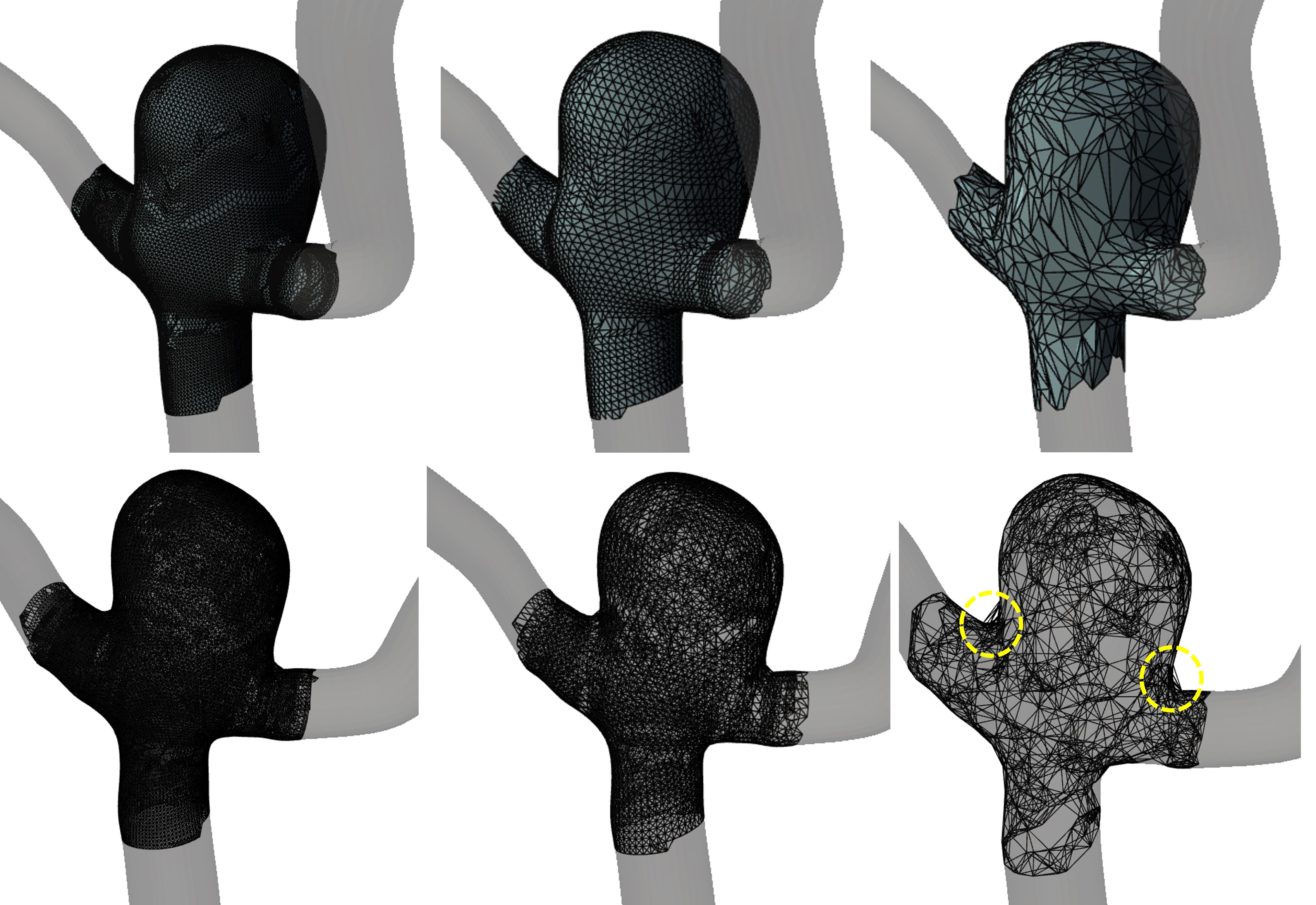}}
\caption{\textbf{Top:} GHD-deformed mesh and the down-sampled versions. \textbf{Bottom:} KNN-constructed graphs where false connections are observed in highlighted regions.}
\label{fig3}
\end{figure}

\begin{figure*}[!t]
    \includegraphics[width=1\textwidth]{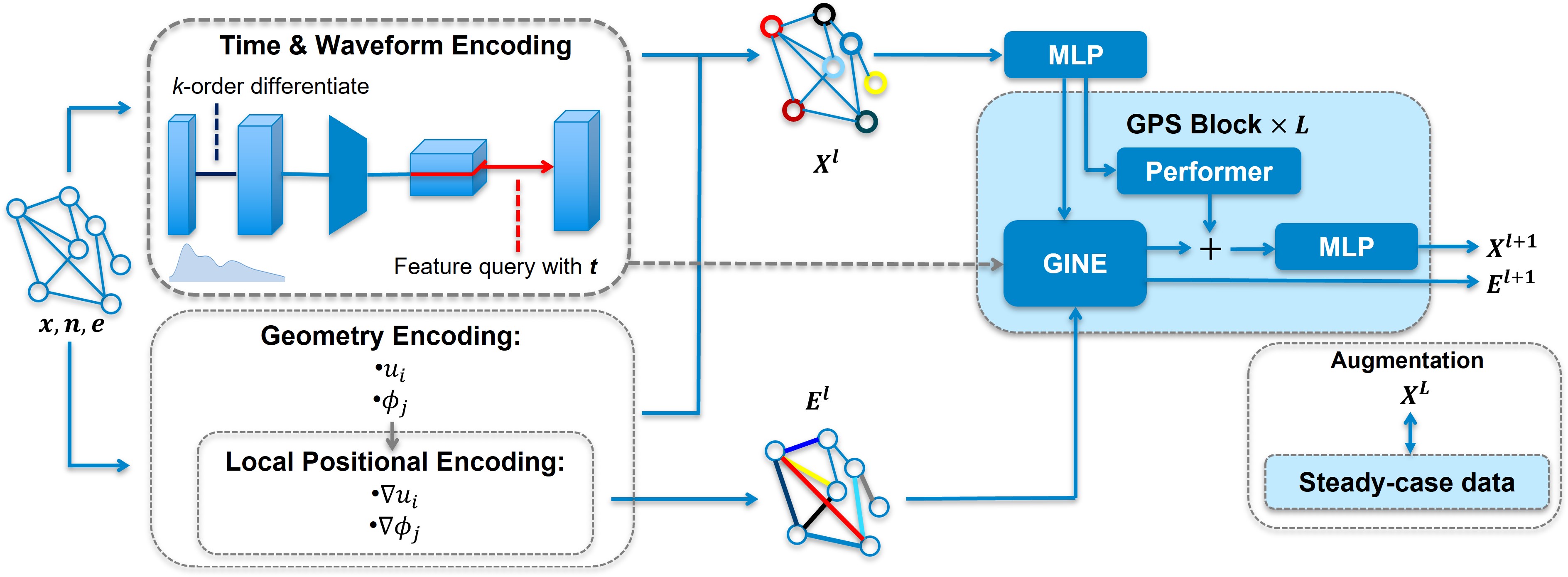}
    \caption{\textbf{Network schematic of our Graph Transformer.}}
    \label{network_schematic}
\end{figure*}

We take the surface mesh of each aneurysm as the representation of their morphology. Geometries in the AneuG-Flow dataset are created by a deep learning generative model utilizing the Graph Harmonic Deformation encoding \cite{luoGHD2024}. GHD creates a series of Graph Fourier Basis (shape modes) $U:=[u_1, ...,u_n]$ from a canonical/template shape by solving the eigenvectors of its surface mesh's cotangent graph Laplacian matrix $L_c$:

\begin{equation}
L_c \cdot u_i = \lambda_i\, u_i, \qquad 0 < i \leq N .
\end{equation}
Here, shape modes with large eigenvalues correspond to low-frequency behaviors on the graph, and vice versa (as shown in Fig.~\ref{fig2}). $N$ is the node number of the canonical shape, leading to $N$ eigenvectors in total. More technical details can be found in \cite{luoGHD2024}. In the dataset, each shape is accompanied by a checkpoint file containing a set of scalar tuples $\phi_i=(\phi_{ix},\phi_{iy},\phi_{iz})$ refered to as GHD tokens. The watertight surface mesh of each shape can then be reconstructed with the GHD tokens as:

\begin{equation}
    \mathcal{M} \bigg( \mathcal{V} = \sum_{i=1}^n U_i \cdot \phi_i, \mathcal{F}_c \bigg)
\end{equation}
where $\mathcal{V}_c$ and $\mathcal{F}_c$ denote the mesh node coordinates and triangle faces of the canonical shape. A few examples of deformed shapes are visualized in Fig.~\ref{fig3}. By using GHD reconstruction instead of directly using the surface graph of the CFD volume meshes, we obtain a uniform graph structure and achieve downsampling that retains high fidelity across all shapes in the dataset. Compared with the dynamic Farthest Point Sampling plus KNN approach used in existing works such as \cite{sukLaBGATr2024}, our approach simplifies the workflow and reduces computational cost. We only need to pre-compute and store a single set of graph down-sampling and up-sampling operations before training begins, with no additional computation required on the fly. More importantly, we adopt the topology-preserving method from \cite{luoGHD2024} instead of directly constructing the down-sampled graph using KNN \cite{sukLaBGATr2024}. As shown in Fig.~\ref{fig3}, this prevents false connections being created between points that are geometrically close but topologically distant. 
When using our model for a user-specific shape, a fitting process is needed to acquire the set of $\phi_i$. This generally takes a matter of minutes.

\subsection{Geometry Encoding}
\label{geometry_encoding}
As mentioned earlier, U-Net and Transformer are two mostly used modules for geometry feature extraction. The U-Net aggregates information through graph convolution and pooling, and form a latent vector representing the geometry at the bottleneck. For our transformer, we follow the GPS Graph Transformer \cite{rampasekGraphGPS2023} recipe. As shown in Table.~\ref{feature_summary}, for positional encodings, We have the GHD eigenvectors (shape modes) $u_i$ and independent eigenvectors $\phi_j$ of each mesh's cotangent Laplacian eigenvectors. We also compute their spatial gradients as $\nabla u_i$ and $\nabla\phi_j$. In this work, we truncate the total number of $i$ and $j$ to 8 and 16 independently. Corresponding eigenvalues are included as structural encoding features. Unlike most graph transformer works, We do not include anything related to the random walk matrix. This is because we are using triangle meshes and the random walk matrix would carry very limited information in regions not connected to the mesh boundary. Instead, we simply include the one-hot encoding features informing the network of the mesh's open boundaries. Node and edge features are summarized in Table.~\ref{tab2}.

\begin{table}
\caption{Feature summary of our Graph Transformer}
\label{tab2}
\setlength{\tabcolsep}{3pt}
\begin{tabular}{|p{45pt}|p{45pt}|p{125pt}|}
\hline
Symbol& Feature type& Quantity \\
\hline
\multicolumn{3}{|c|}{Positional encoding} \\ 
\hline
$u_i$ & node & GHD shape modes \\
\hline
$\phi_j$ & node & Cotangent Laplacian eigenvectors \\
\hline
$\nabla u_i$ & node & Spatial gradients of $u_i$ \\
\hline
$\nabla\phi_j$ & node & Spatial gradients of $\phi_j$\\
\hline
\multicolumn{3}{|c|}{Structural encoding} \\ 
\hline
$\lambda_i$ & node & GHD eigenvalues \\
\hline
$\mu_j$ & node & Cotangent Laplacian eigenvalues \\
\hline
\multicolumn{3}{|c|}{Other} \\ 
\hline
$\mathbf{e}$ & node & One-hot encoding of node type\\
\hline
$\mathbf{x}$ & node & Euclidean coordinates\\
\hline
$\mathbf{n}$ & node & Surface normals\\
\hline
\multicolumn{3}{p{251pt}}{Feature summary of graph Transformer input.}
\end{tabular}
\label{feature_summary}
\end{table}

\subsection{Time and Waveform Encoding}
\label{time_waveform_encoding}
Since variations in mass flow at the inlet directly determine the flow flux, it is necessary to extract informative features from the waveform signal. Moreover, certain behaviors of the WSS may correlate with higher-order dynamics of the waveform. For example, rapid acceleration in flow (i.e., the derivative of velocity) often leads to sharper WSS changes. Therefore, we compute the first $k$ order derivatives of the signal and use them as inputs to a 1D convolutional U-Net encoder. During training, we slice the features to the corresponding time frame and inject them into the network. For our Graph Transformer (Fig.~\ref{network_schematic}), these waveform features are treated as structural encodings and are injected repeatedly into every GPS block. For the Graph U-Net, we simply append them at the bottleneck layer.

\subsection{Augmentation with Cheap Steady Data}
\label{steady_augmentation}
For CFD simulations, steady cases are significantly cheaper compared to transient ones. For instance, one transient CFD simulation in the AneuG-Flow dataset took around 2,000 time steps. This means the same computational resource for one transient case is equivalent to thousands of steady cases. This ratio is even larger if parallel computational is taken into consideration. Therefore, we propose using the unpaired cheap steady data as augmentation. In practice, we simply mask the time \& waveform features of these data and train our baseline models with both steady and transient cases.

\subsection{Comparative Baselines}
To benchmark the performance of our proposed Graph Transformer, we compare several competitive baselines, which can be categorized into three families: per-snapshot surrogates, sequence surrogates, and spectral modal surrogates.

\textbf{Per-snapshot Surrogates} regress WSS at a single time step and are evaluated by aggregating predictions across the full cardiac cycle. First, we implement Graph U-Net \cite{gaoGraphUNets2019} to isolate the efficacy of Transformer-based global attention. The network follows the multi-scale encoder-decoder structure and uses chebyshev spectral graph convolutional layers (ChebConv) and pre-computed down/up-sampling indices. Second, we adapt LaB-GATr \cite{sukLaBGATr2024} and LaB-VATr \cite{sukvatr2024} baselines, which represent recent transformer-based surrogates for geometric hemodynamics prediction. LaB-GATr \cite{rygiel2025} was reported as a state-of-the-art transient WSS surrogate in synthetic abdominal aortic aneurysm models. We implement these baselines on the same standardized surface meshes using only coordinates and surface normals as geometric descriptors.

\textbf{Sequence Surrogate} predicts the entire cardiac-cycle WSS in a single forward pass rather than regressing each time step independently. This baseline uses the same Graph U-Net \cite{gaoGraphUNets2019} geometry encoder to extract node-wise spatial features, which are fused with inflow waveform features using cross-attention. The resulting joint features are then decoded by several MLP layers to produce the full WSS sequence across the cardiac cycle. We also incorporate a steady WSS prior (predicted by a separate model trained on the 14,000 steady cases) using Feature-wise Linear Modulation (FiLM) to test whether the cheap steady augmentation provides a useful anchor for transient prediction.

\textbf{Modal Surrogate} is a spectral domain baseline that predicts WSS in a truncated modal space defined by the GHD shape modes. Specifically, given the GHD eigenbasis $U=[u_1,\dots,u_N]$ (Section~\ref{shape_preprocessing}), we keep the first $K=512$ modes and project the transient WSS field onto this subspace:
\begin{equation}
\mathbf{c}(t)=U_{1:K}^{\top}\mathbf{w}(t), \qquad 
\hat{\mathbf{w}}(t)=U_{1:K}\hat{\mathbf{c}}(t),
\label{eq:modal_projection}
\end{equation}
where $\mathbf{w}(t)\in\mathbb{R}^{N\times 3}$ denotes the WSS field at time $t$, and $\mathbf{c}(t)\in\mathbb{R}^{K\times 3}$ denotes its spatial modal coefficients. The model maps the GHD coefficients directly to the WSS spectral coefficients, and the full-resolution WSS sequence is reconstructed by inverse projection. This baseline provides a reduced-order reference that helps diagnose high-frequency artifacts observed in other surrogates.

\subsection{Implementation Details}
The dataset was divided into training and testing sets with a 9:1 ratio. Coordinates were normalized using the dataset-wide mean and standard deviation. To augment the transient simulation data with steady cases, we dynamically sampled from both transient snapshots and the steady dataset, and the temporal features were masked to zero for steady samples. For optimization, we used AdamW optimizer with an initial learning rate of $3\times10^{-4}$ and a weight decay of $1\times10^{-4}$. We also implemented a StepLR scheduler with a step size of 50 and a decay factor of 0.75. The Graph Transformer model was trained for 250 epochs with a batch size of 10 per GPU. We saved the checkpoint with the minimum Mean Squared Error (MSE). All experiments were conducted on NVIDIA A40 (48GB) GPUs, using PyTorch 2.4 and PyTorch Geometric.

\subsection{Metrics}
We evaluate model performance with Mean Squared Error (MSE) for the predicted WSS vector across the aneurysm surface mesh nodes:
\begin{equation}
\mathrm{MSE}
=
\mathbb{E}_{\Omega}
\;
\mathbb{E}_{\mathbf{x} \sim \Omega}
\left[
    \left\lVert
        \mathbf{w}^{\text{pred}}(\mathbf{x} \mid \Omega)
        -
        \mathbf{w}^{\text{true}}(\mathbf{x} \mid \Omega)
    \right\rVert_2^{2}
\right],
\end{equation}
where $\Omega$ represents a set of aneurysm surface manifolds in the test dataset. $\mathbf{x}$ and $\mathbf{w}$ are Euclidean coordinates and WSS vectors, respectively. 

To provide a more interpretable evaluation, we report two relative $\ell_2$ metrics. We define the conventional relative $\ell_2$ error as
\begin{equation}
\mathrm{rL2}
=
\mathbb{E}_{\Omega,t}
\;
\left[
\frac{
\left\lVert 
\mathbf{w}^{\text{pred}}(t) - \mathbf{w}^{\text{true}}(t)
\right\rVert_{F}
}{
\left\lVert 
\mathbf{w}^{\text{true}}(t)
\right\rVert_{F}
}
\right].
\end{equation}
However, directly normalizing by $\|\mathbf{w}^{\text{true}}(t)\|_F$ can be unstable for transient WSS. Large portions of the surface region have low WSS magnitude, and $\|\mathbf{w}^{\text{true}}(t)\|_F$ can be close to zero. As a result, $\mathrm{rL2}$ may be dominated by low-magnitude regions and can over-amplify errors. To address this issue, we introduce a modified relative $\ell_2$ metric $\mathrm{rL2^*}$, which normalizes the error using a per-case maximum WSS magnitude:
\begin{equation}
\mathrm{rL2^*}
=
\mathbb{E}_{\Omega,t}
\;
\left[
\frac{
\left\lVert 
\mathbf{w}^{\text{pred}}(t) - \mathbf{w}^{\text{true}}(t)
\right\rVert_{F}
}{
\max\left\lVert \mathbf{w}^{\text{true}}(t)\right\rVert_{F}
}
\right].
\end{equation}
The maximum WSS magnitude scale over the cardiac cycle for each geometry prevents extremely small WSS from dominating the relative error, yielding a more stable and visually consistent assessment.

Driven by the same motivation, we also employ the Structural Similarity Index Measure (SSIM). SSIM cannot be applied directly to our predictions, as the WSS field is defined on an irregular 3D manifold instead of a regular image grid.  We therefore transform the surface node features as a 2D image through reshaping and padding before computing SSIM. However this does not respect the true spatial relationships on the mesh, making the metric less interpretable. To preserve spatial structure, we thus introduce another SSIM measure that is more visually meaningful evaluation through mesh rendering ($SSIM_r$). We treat the WSS magnitude as a surface texture, and render each aneurysm mesh into a set of 2D rendered images from $N$ uniformly spaced viewpoints ($N=6$ in our experiments). The final SSIM score is obtained by averaging the SSIM values from all rendered views:
\begin{equation}
\mathrm{SSIM_r}
=
\frac{1}{N}
\sum_{i=1}^{N}
\mathrm{SSIM}\big(I^{\text{pred}}_{i},\, I^{\text{true}}_{i}\big),
\end{equation}
where $I^{\text{pred}}_{i}$ and $I^{\text{true}}_{i}$ denote the predicted and ground-truth WSS-magnitude renderings from viewpoint $i$, respectively.

\section{Results}

\subsection{Performance Comparison}
We compare our proposed Graph Transformer against per-snapshot, sequence, and spectral baselines, with and without steady data augmentation ($\dag$). Table.~\ref{tab_baseline_comparison} summarizes the quantitative results.Among the augmentated per-snapshot baselines, our model ($T_1^*$) achieves the best performance with an MSE of 0.179 and $rL2^*$ of 2.84\%. It outperforms the per-snapshot U-Net ($U_1^*$; MSE 0.215 and $rL2^*$ 3.20\%), LaB-GATr \cite{sukLaBGATr2024} ($G_1^*$; MSE 0.359 and $rL2^*$ 4.55\%), and LaB-VATr \cite{sukvatr2024} ($V_1^*$; MSE 0.204 and $rL2^*$ 3.40\%). This indicates that our GHD-based embedding and global self-attention are more effective for geometric hemodynamic mappings than the competing baselines.

The results also show that it is more difficult to regress the full spatiotemporal field in a single forward pass. The sequence U-Net surrogate ($Q_1^*$) performs poorly compared to the Transformer method ($rL2^*$ 4.36\% vs 2.84\%). Similarly, the spectral surrogate ($S_1^*$) achieves a competitive $rL2^*$ of 3.32\%, but remains below the Transformer method. In additional, both sequence and spectral methods show a high dependency on steady-data augmentation, and their performance degrades significantly without it. This suggests that steady supervision provides a stabling constraint. We quantify this effect under limited transient training data in Section~\ref{steady_augment}.

\begin{figure}[!t]
 \centerline{\includegraphics[width=\columnwidth]{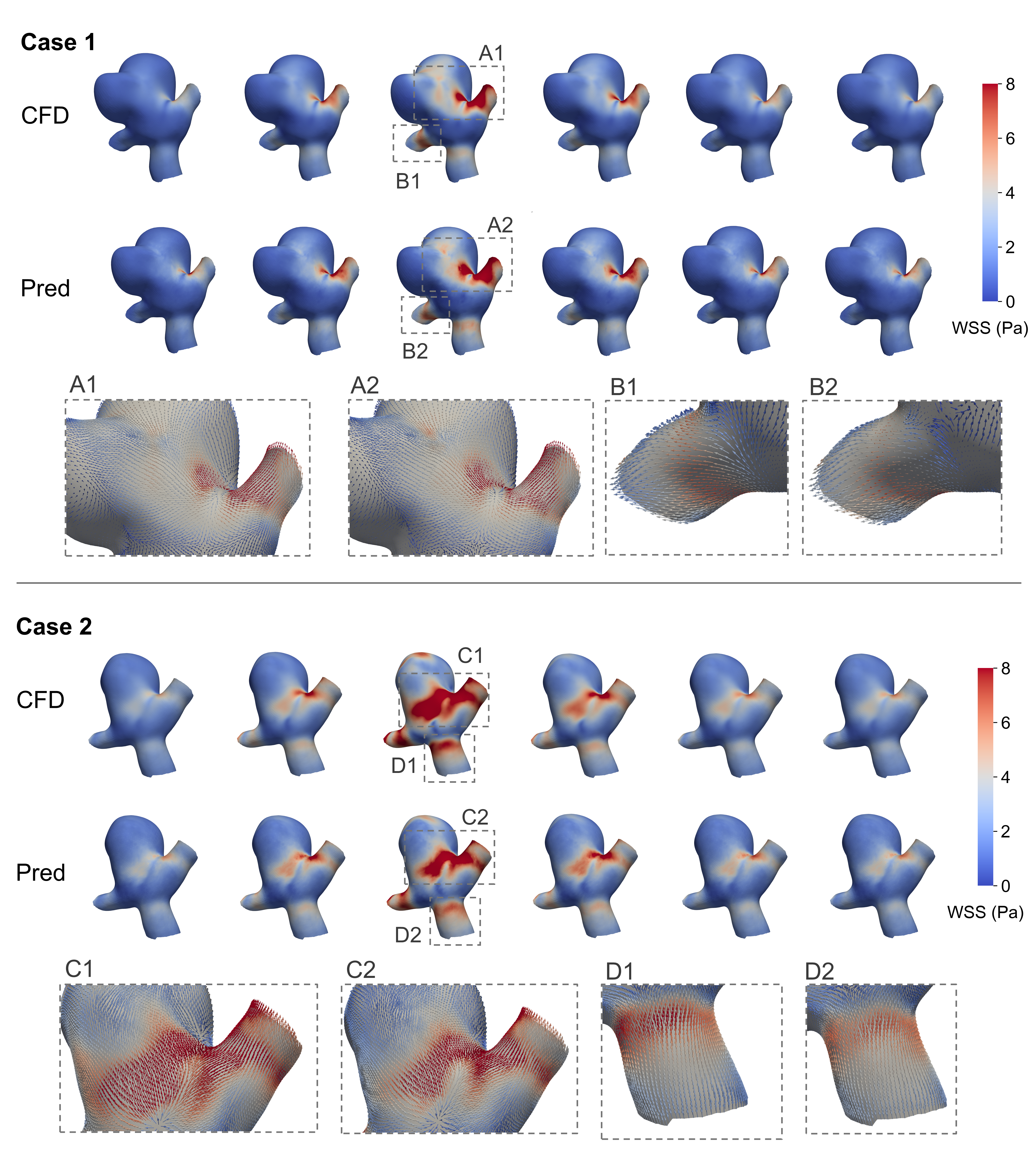}}
 \caption{Qualitative comparison of CFD and DL-predicted transient WSS field.}
 \label{fig4}
\end{figure}

Fig.~\ref{fig4} presents the qualitative comparison between CFD ground truth and predicted WSS fields for two representative cases across the cardiac cycle. The proposed Transformer baseline accurately captures the location and extent of high-WSS regions along the aneurysm neck and dome, and the peak systole insets show good alignment in WSS direction, while noticeable differences remain. We observe noisy transitions in some predictions, particularly for the per-snapshot and sequence baselines, whereas the spectral surrogate produces smoother fields. These behaviors are analyzed in the Discussion.

\begin{table*}[!t]
  \centering
  \begin{threeparttable}
  \caption{Baseline Performance Comparison}
  \label{tab_baseline_comparison}
  \begin{tabular}{cccccccccc}
    \toprule
    Index &Method &$\dag$ &Train-set size &MSE $\downarrow$ &SSIM $\uparrow$ &SSIM$_r$ $\uparrow$ &$rL2$ ($\%$) $\downarrow$ &$rL2^*$ ($\%$) $\downarrow$ & $\Delta rL2^* ($\%$)$\\
    \midrule
    $U_1$ & U-Net &\ding{55} &520 &0.336 &0.967 &0.904 &36.88 &4.05 &-\\
    $U_1^*$ & U-Net &\checkmark &520 &0.215 &0.981 &0.913 &29.51 &3.20 &-0.85 \\
    $G_1$ & LaB-GATr \cite{sukLaBGATr2024} &\ding{55} &720 &0.378 &0.919 &0.976 &38.44 &4.79 &-\\
    $G_1^*$ & LaB-GATr \cite{sukLaBGATr2024} &\checkmark &720 &0.359 &0.924 &0.977 &36.78 &4.55 &-0.24\\
    $V_1$ & LaB-VATr \cite{sukvatr2024} &\ding{55} &720 &0.255 &0.924 &0.987 &32.13 &3.87 &-\\
    $V_1^*$ & LaB-VATr \cite{sukvatr2024} &\checkmark &720 &0.204 &0.932 &0.989 &28.04 &3.40 &-0.47\\
    $Q_1$ & U-Net (Sequence) &\ding{55} &720 &0.524 &0.776 &0.976 &36.90 &6.03 &-\\
    $Q_1^*$ & U-Net (Sequence) &\checkmark &720 &0.272 &0.938 &0.984 &25.83 &4.36 &-1.67 \\
    $S_1$ & Spectral Surrogate &\ding{55} &720 &0.639 &0.723 &0.971 &68.22 &6.23 &-\\
    $S_1^*$ & Spectral Surrogate &\checkmark &720 &0.183 &0.942 &0.987 &28.82 &3.32 &-2.91\\
    $T_1$ & Ours (Transformer) &\ding{55} &520 &0.262 &0.973 &0.961 &32.51 &3.45 &-\\
    $T_1^*$ & \textbf{Ours (Transformer)} &\checkmark &520 &\textbf{0.179} &\textbf{0.982} &0.973 &26.98 &\textbf{2.84} &-0.61\\
  \bottomrule
\end{tabular}
\begin{tablenotes}
\footnotesize
\item[$\dag$] Indicates if steady-data augmentation is used; SSIM$_r$ is the rendered-SSIM; $\Delta rL2^*$ is the improvement of $rL2^*$ over the non-augmented baseline. $\uparrow$: larger value is better; $\downarrow$: vice versa. 
\end{tablenotes}
\end{threeparttable}
\end{table*}

\begin{table*}[!t]
  \centering
  \begin{threeparttable}
  \caption{Data Efficiency Analysis}
  \label{tab_steady_augmentation}
  \begin{tabular}{cccccccccc}
    \toprule
    Index &Backbone &$\dag$ &Train-set size &MSE $\downarrow$ &SSIM $\uparrow$ &SSIM$_r$ $\uparrow$ &$rL2$ ($\%$) $\downarrow$ &$rL2^*$ ($\%$) $\downarrow$ &$\Delta rL2^* ($\%$)$\\
    \midrule
    $U_1$ & U-Net& \ding{55} &520 &0.336 &0.967 &0.904 &36.88 &4.05 &-\\
    $U_1^*$ & U-Net& \checkmark &520 &0.215 &0.981 &0.913 &29.51 &3.20 &-0.85\\
    $U_2$ & U-Net& \ding{55} &250 &0.488 &0.951 &0.901 &44.37 &4.83 &-\\
    $U_2^*$ & U-Net& \checkmark &250 &0.272 &0.967 &0.921 &33.09 &3.62 &-1.21\\
    $U_3$ & U-Net& \ding{55} &100 &0.923 &0.899 &0.847 &60.95 &6.56 &-\\
    $U_3^*$ & U-Net& \checkmark &100 &0.306 &0.967 &0.901 &35.14 &3.85 &-2.71\\
    $U_4$ & U-Net& \ding{55} &50 &2.189 &0.772 &0.884 &93.8 &9.47 &-\\
    $U_4^*$ & U-Net& \checkmark &50 &0.306 &0.919 &0.908 &44.73 &5.59 &-3.88\\
  \bottomrule
\end{tabular}
\begin{tablenotes}
\footnotesize
\item[$\dag$] Indicates if steady-data augmentation is used; SSIM$_r$ is the rendered-SSIM; $\Delta rL2^* ($\%$)$ is the improvement of $rL2^*$ over the non-augmented baseline. $\uparrow$: larger value is better; $\downarrow$: vice versa. 
\end{tablenotes}
\end{threeparttable}
\end{table*}

\subsection{Derived Hemodynamic Metrics}
To further validate our model, we derived TAWSS, OSI, and RRT from the predicted transient WSS fields. These are defined as:

\begin{equation} \mathrm{TAWSS} = \frac{1}{T} \int_{0}^{T} |\mathbf{w}(t)|  dt, \end{equation}
\begin{equation} \mathrm{OSI} = \frac{1}{2} \left( 1 - \frac{|\int_{0}^{T} \mathbf{w}(t) , dt|}{\int_{0}^{T} |\mathbf{w}(t)| , dt} \right) \end{equation}
\begin{equation} \mathrm{RRT} = \frac{1}{(1 - 2 \cdot \mathrm{OSI}) \cdot \mathrm{TAWSS}} \end{equation}

Table.~\ref{tab:derived_metrics} reports the quantitative results of these indices. TAWSS shows good agreement with CFD. The overall MSE is 0.281 with an $rL2^*$ of 3.81\% and an SSIM of 0.80. Similarly, RRT is predicted with high accuracy, with an MSE of 0.127, the lowest $rL2^*$ among the three indices (2.62\%), and a high SSIM of 0.96. In contrast, OSI shows the largest relative error (68.3\%), while its $rL2^*$ remains moderate at 5.34\%. This discrepancy is expected because in this dataset, OSI values are generally small ($<0.1$) for most of the aneurysm surface, and high oscillations are only in a small region. As a result, the denominator in the $rL2$ is often close to zero, which leads to high percentage error. This is consistent with the small MSE (0.000376). When the error is normalized with the average-based metric $rL2^*$, it drops to 5.34\%.

\begin{figure}[!t]
 \centerline{\includegraphics[width=\columnwidth]{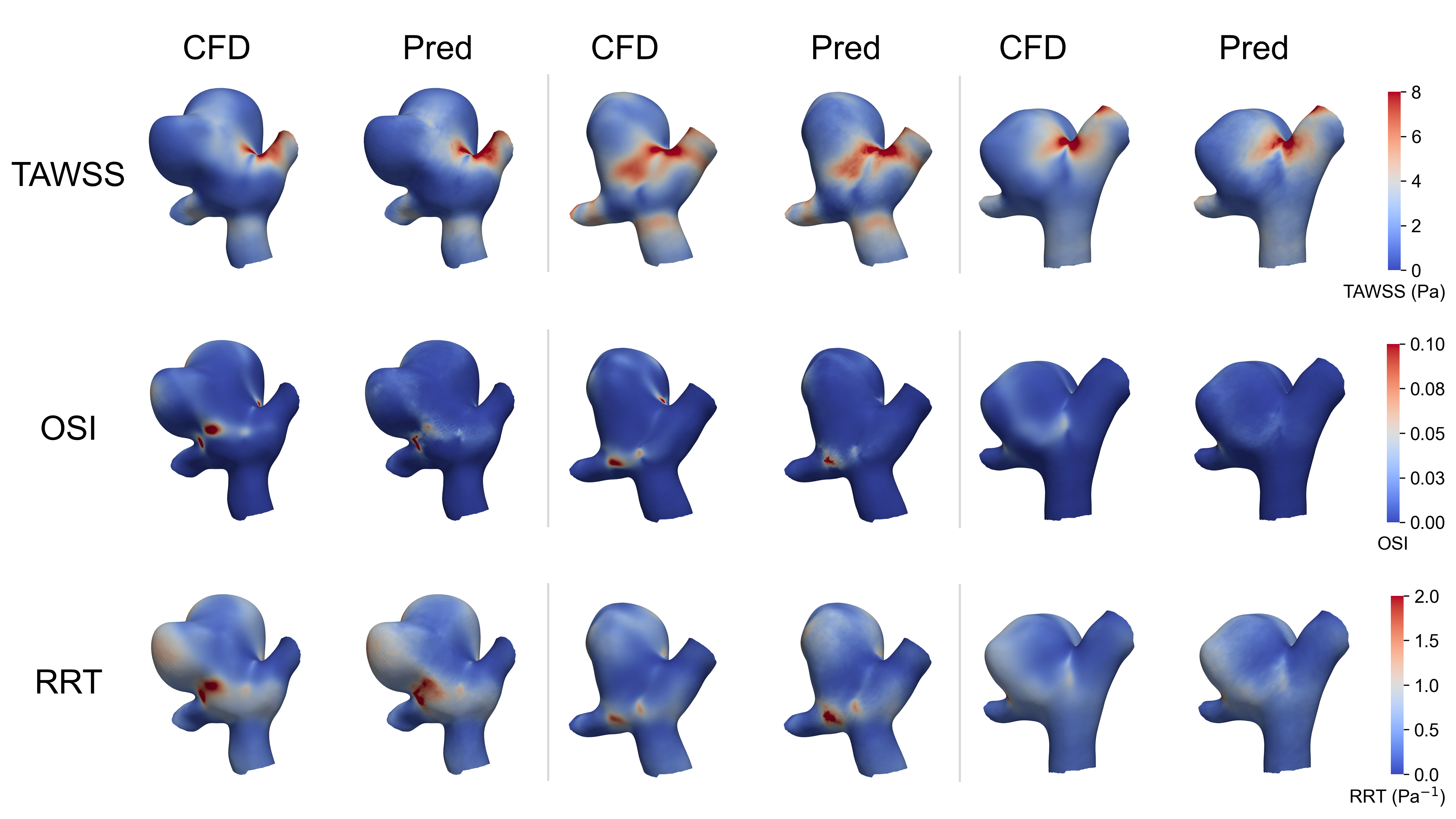}}
 \caption{Qualitative comparison of CFD and DL-derived TAWSS, OSI, and RRT fields.}
 \label{fig_tawss_osi_rrt}
\end{figure}

Figure.\ref{fig_tawss_osi_rrt} illustrates TAWSS, OSI, and RRT for three representative aneurysms. The derived TAWSS maps accurately capture the location of high-shear zones near the aneurysm neck and branch vessel. The RRT maps identify regions of flow stagnation in the aneurysm neck. Despite the numerical sensitivity of OSI, the distributions of OSI fields are captured, although the derived fields are smoother compared to the CFD ground truth.

\begin{table}[t]
    \centering
    \caption{Quantitative Evaluation of Derived Hemodynamic Metrics}
    \label{tab:derived_metrics}
    \begin{tabular}{l c c c c}
    \toprule
    Metric & MSE & SSIM & $rL2$ ($\%$) & $rL2^*$ ($\%$) \\
    \midrule
    TAWSS & 0.281 & 0.800 & 13.54 & 3.81 \\
    OSI & 3.76e-4 & 0.741 & 68.35 & 5.34 \\
    RRT & 0.127 & 0.961 & 17.31 & 2.62 \\
    \bottomrule
\end{tabular}
\end{table}

\subsection{Ablation on Network Architecture}
To identify key components in the network design that contribute most to the performance, we conduct an ablation experiment on the positional and structural encoding of our Graph Transformer. Specifically, we remove the node and edge features related to GHD and cotangent graph Laplacian respectively. When both are deactivated, the Transformer relies solely on the node coordinates and surface normals to act as positional encoding. This is similar to the design of LaB-GATr except the surface mesh connection is retained in the data. The results are presented in Table.~\ref{tab_gps_ablation}. It is observed that both the GHD-related and case-specific cotangent Laplacian features positively contribute to the overall network performance. When both feature types are deactivated, the architecture reduces to a LaB-GATr–like model. However, the overall performance is still slightly better compared to LaB-GATr and LaB-VATr. This behavior is most likely due to the node–edge feature fusion strategy employed in the Graph GPS architecture.

\begin{table}[t]
    \centering
    \caption{Ablation on GHD and cotangent Laplacian  feature encoding}
    \label{tab_gps_ablation}
    \begin{tabular}{l c c c c}
    \toprule
    GHD & Cotangent Laplacian & SSIM$\uparrow$ & $rL2$ ($\%$)$\downarrow$ & $rL2^*$ ($\%$)$\downarrow$ \\
    \midrule
    \ding{55} & \ding{55} & 0.973 & 31.86 & 3.34 \\
    \checkmark & \ding{55} & 0.974 & 28.28 & 2.97 \\
    \checkmark & \checkmark & 0.982 & 26.98 & 2.84 \\
    \bottomrule
\end{tabular}
\end{table}

\subsection{Ablation with Steady Augmentation}
\label{steady_augment}
As mentioned earlier, we include inexpensive steady‐state CFD data during training to compensate for the relatively small size of the pulsatile dataset. In this section, we present an ablation study to examine how effective this strategy is. A quantitative summary of the model performances is provided in Table.~\ref{tab_steady_augmentation}.

We evaluated both our per-snapshot U-Net baseline under different transient training set sizes, with and without steady-data augmentation. Across all settings, steady-data augmentation improves model performance, and this benefit becomes more obvious as the available pulsatile training data decreases. For example, when the dataset size is reduced to 100 cases, steady-data augmentation lowers the $rL2^*$ error from 6.56\% to 3.85\% and increases the SSIM from 0.899 to 0.967 (Table.~\ref{tab_steady_augmentation}). Rendered SSIM$_r$ experiences the same pattern from 0.847 to 0.901. From an engineering perspective, this improvement matters for two key reasons.

\noindent \textbf{1). Quality threshold:} An SSIM below 0.90 generally indicates lower-quality reconstruction. Thus, the augmentation can provide a step-wise increase in reconstruction fidelity when the dataset is small.

\noindent \textbf{2). Resource limit:} For CFD simulations at this resolution, dataset of 100 transient cases is already considered substantial due to the computational expense. Each pulsatile simulation in AneuG-Flow dataset took at least 12 hours on a 64-core Intel Ice Lake Xeon Platinum 8358 system with 96 GB RAM. In contrast, tens of thousands of steady simulations can be completed in the same amount of time, making them a highly cost-effective source of additional information.

\begin{figure*}[!t]
    \includegraphics[width=1\textwidth]{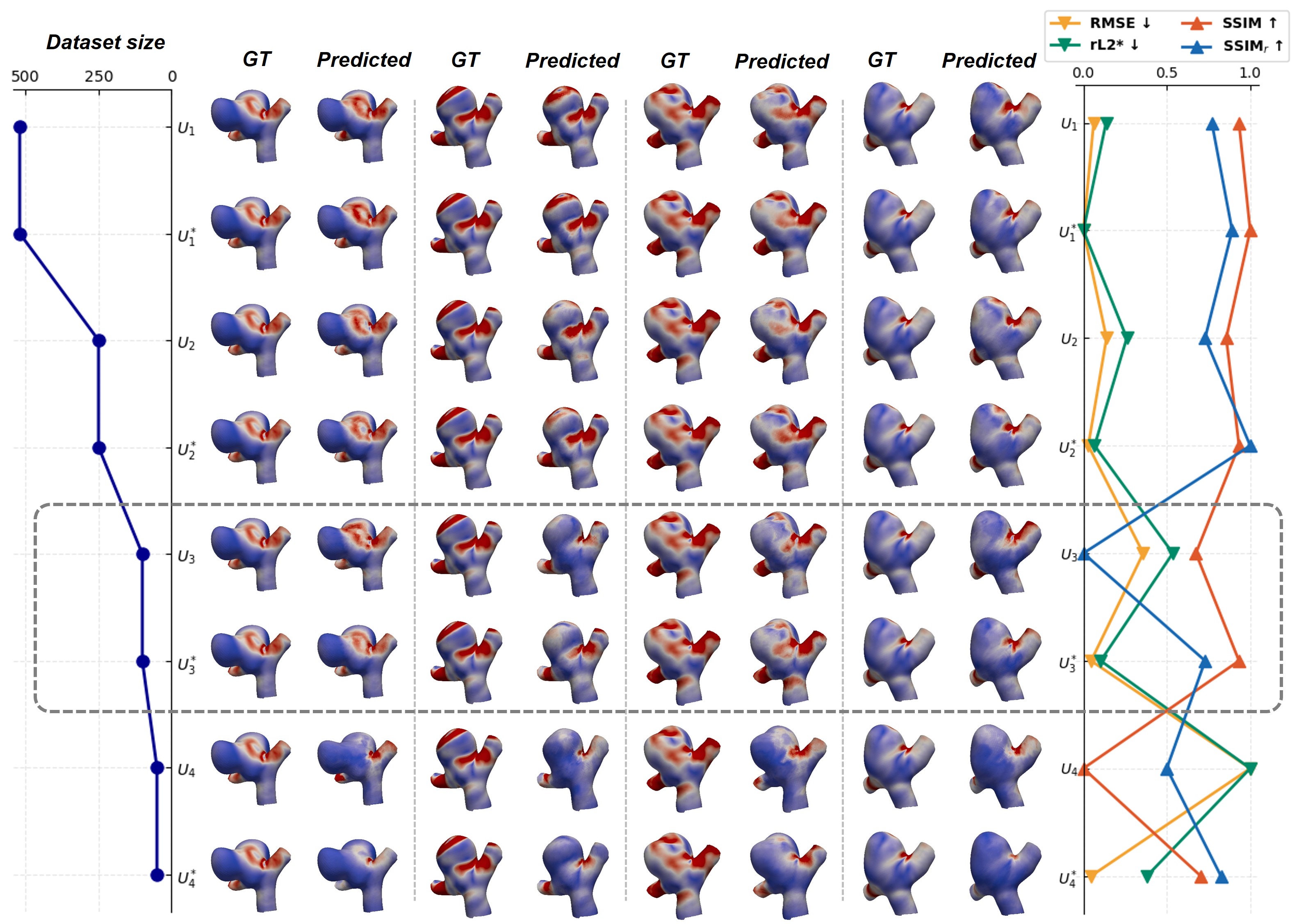}
    \caption{\textbf{Ablation on Graph U-Nets w/o steady-data augmentation. Left:} Training dataset size. \textbf{Middle:} Ground truth WSS maps versus predicted maps. \textbf{Right:} Performance metrics each normalized to range from zero to one.}
    \label{fig_steady_augmentation}
\end{figure*}

To better illustrate the effect, we randomly selected several aneurysms from the test set and visualized the predicted WSS field at the waveform-peak frame (Fig.~\ref{fig_steady_augmentation}). Even when the numerical metrics are similar, the visual improvement in WSS distribution—particularly over the aneurysm pouch—is noticeable (e.g., $U_1$ vs $U_1^*$ in (f) and (h)). The effect is even clearer when comparing $U_3$ and $U_3^*$, both trained on 100 pulsatile cases. Without augmentation ($U_3$), the predicted WSS near the impingement point around the neck and at the top of the dome deviates from the ground truth.

\begin{figure*}[!t]
    \includegraphics[width=1\textwidth]{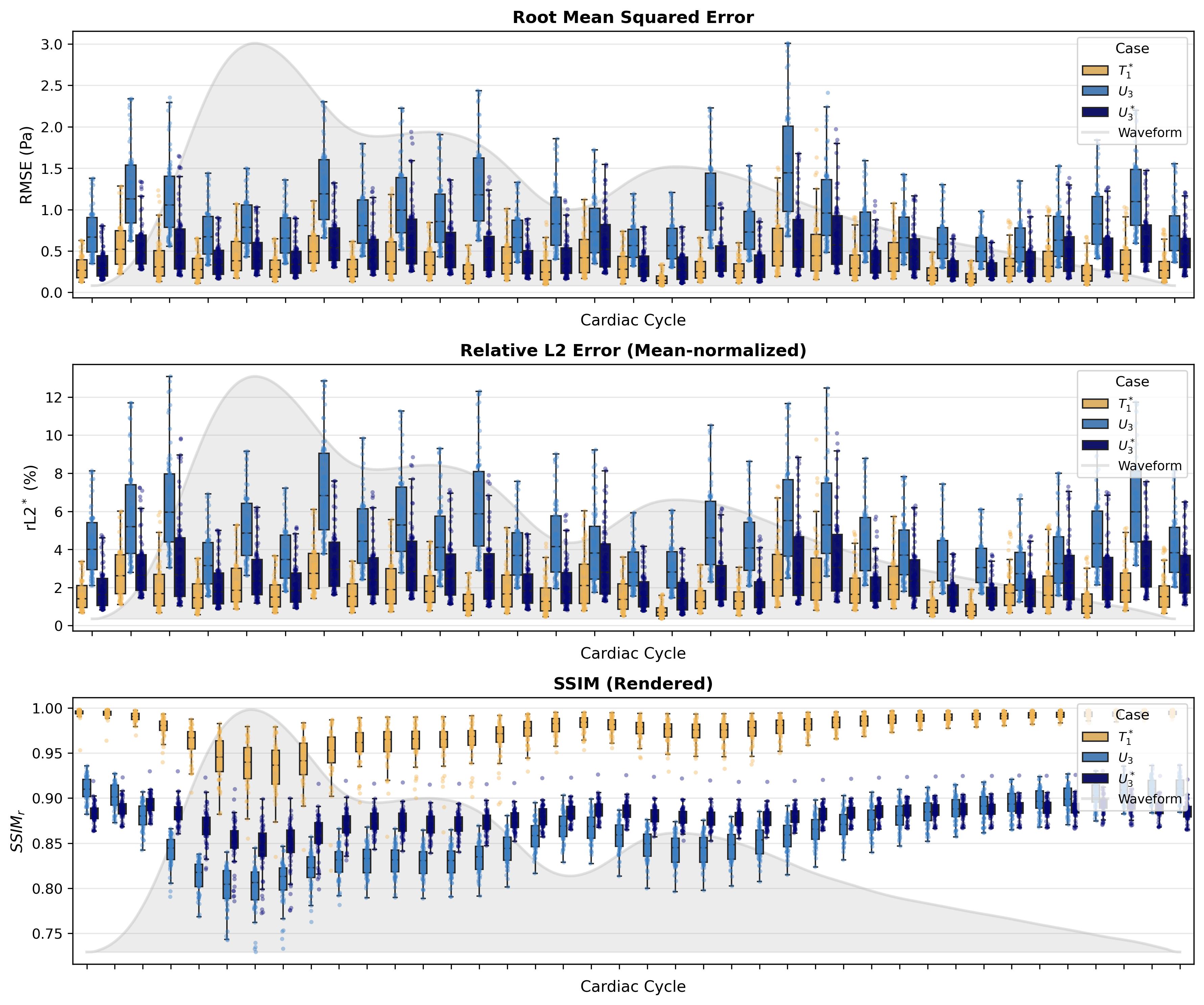}
    \caption{\textbf{Model performance throughout the cardiac cycle.}}
    \label{fig_ablation_metrics}
\end{figure*}

We further analyze model performances over the cardiac cycle for representative cases $T_1^*$, $U_3$, and $U_3^*$ in Fig.~\ref{fig_ablation_metrics}. Comparing $U_3$ and $U_3^*$, we see that steady-data augmentation consistently reduces $rL2^*$ across almost all frames. All models exhibit larger errors during the late-diastolic phase, likely because the WSS magnitude in these frames is much lower than the systole stage, making them less emphasized during training.

For SSIM$_r$, below-average performance occurs around the peak flow moment. This is probably due to the presence of more complex, unsteady WSS patterns in some geometries at peak systole (see video V1 in Supplementary Material). Nevertheless, the Graph Transformer still provides reasonable predictions, with average SSIM$_r$ values exceeding 0.90.

\section{Discussion}
The study demonstrates that our framework can accurately predict transient WSS fields from aneurysm surface meshes and inlet waveforms. The proposed GHD-enabled Graph Tranformer achieves the best overall performance with a 2.84\% $rL2^*$. The derived hemodynamic metrics, including TAWSS, OSI, and RRT, further validates that the model is expressive enough for downstream computation of these indices on non-idealized aneurysms.

\subsection{Architecture Selection}
The per-snapshot baseline comparison shows that the GHD-enabled Graph Transformer outperforms both Graph U-Net and recent SOTA geometric transformers (LaB-GATr/VATr) \cite{sukLaBGATr2024, sukvatr2024}. We attribute this to two factors: (1) The standardized mesh structure enables consistent, topology-preserving pooling/unpooling indices across the dataset. Compared to FPS and KNN used in previous studies \cite{sukLaBGATr2024, sukvatr2024}, our method better preserves mesh connectivity and avoids false connections as in KNN-constructed graphs (Fig.~\ref{fig3}). (2) GHD positional encodings provide a physically meaningful graph Fourier basis defined on canonical mesh, which allows the network to interpret geometry more effectively than raw coordinates. The ablation study (Table.~\ref{tab_gps_ablation} further supports that adding GHD features improves overall performance.

Among all model families, per-snapshot surrogates outperform both sequence and spectral surrogates. Although sequence models can predict the full WSS fields in a single forward pass, they tend to encourage averaging effect across the cardiac cycle, leading to larger errors around peak systole. This behavior can be attributed to the MSE loss, which encourages the network to converge towards the statistical mean to minimize global residuals, resulting in oversmoothing near peak sytole. In contrast, per-snapshot models with waveform conditioning solve a set of simpler regression problems, which helps preserve localized WSS patterns across the cardiac cycle. The spectral model predicts coefficients in a truncated Laplacian basis and is the most computationally efficient model during training and inference. Since the reconstruction is restricted to low-frequency modes, it produces the smoothest WSS fields and avoids the speckle-like spatial noise observed in other baselines.

\begin{figure}[!t]
 \centerline{\includegraphics[width=\columnwidth]{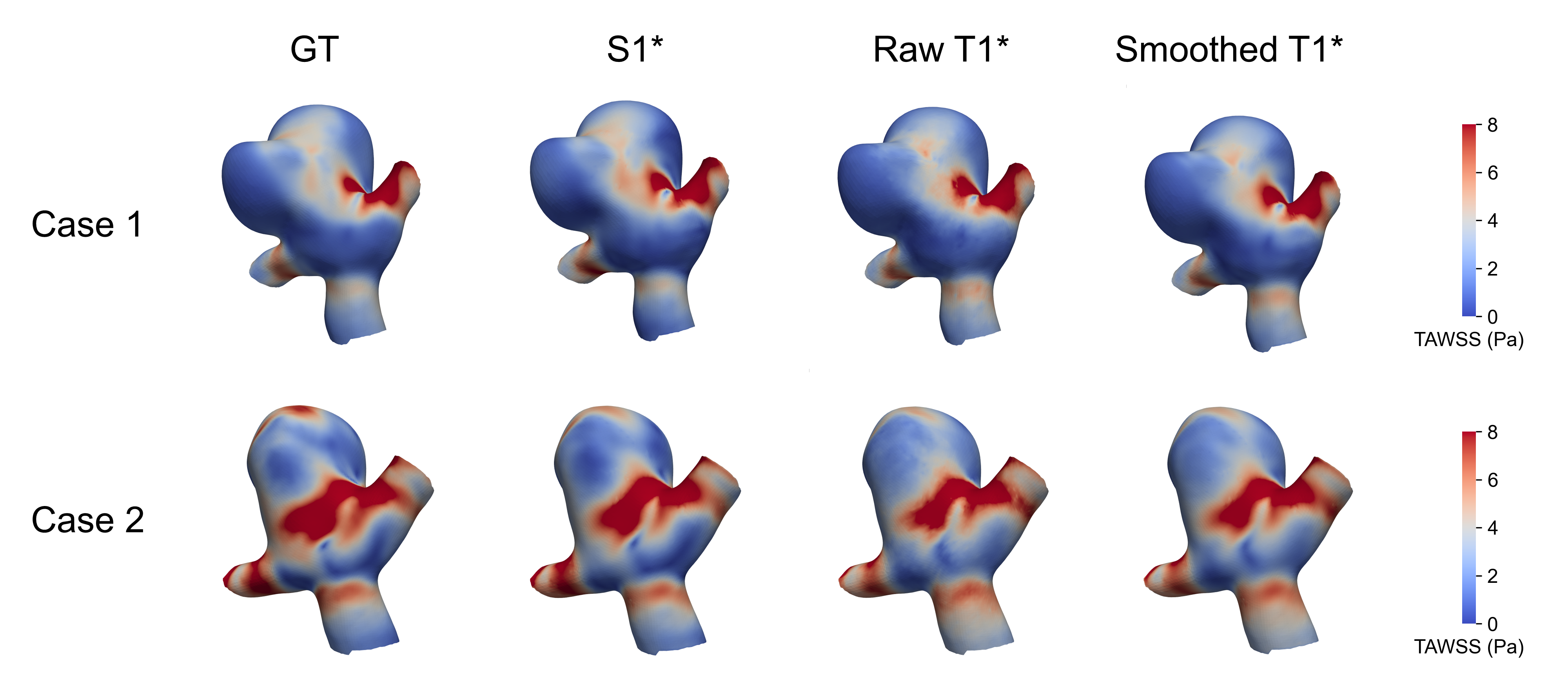}}
 \caption{Effect of Laplacian Low-pass Projection on Spatial Artifacts in Predicted WSS Maps.}
 \label{fig_spatial_smooth}
\end{figure}

We also performed post-hoc experiment to smooth predictions from other models using a truncated Laplacian eigenbasis. This effectively reduces spatial artifacts, but slightly degrades quantitive metrics (around 0.04\% in $rL2^*$). This indicates that the spatial artifacts lives in high-frequency spatial components that contribute limited energy to global norms. While low-pass smoothing removes unstable components, it also suppress true sharp features at high frequency and worsen the metric scores. The qualitive comparison is shown in Fig.~\ref{fig_spatial_smooth}.

\subsection{Steady-data Augmentation}
We evaluated the positive effect of steady-data augmentation in Section \ref{steady_augment}, and found this trick improving model performance for all baselines.
We attribute this improvement primarily to the ability of the large-scale steady dataset to train a robust geometric encoder. Although steady and transient flows differ in temporal dynamics, they share the relationship between geometry and flow, such as the influence of vessel curvature and bifurcations on flow patterns. Training with 14,000 steady-state cases exposes the geometry encoder to a broader diversity of aneurysm morphologies and WSS patterns, resulting in a more generalized geometric representation. Secondly, the transient cases is relatively limited in aneurysm geometries. This leads to higher prediction uncertainty and spatial noise, as shown in Fig.~\ref{fig_steady_augmentation}. The large-scale steady supervision acts as an effective regularizer and produces spatially smoother WSS patterns with fewer high-frequency spatial artifacts.

\subsection{Limitations}
However, our current baseline has several limitations. First, the transient dataset uses a fixed inflow waveform profile across cases, which limits the model’s generalizability to broader physiological conditions with different waveform shapes, amplitudes, and heart rates. Second, the surrogate inherits the modeling assumptions embedded in the CFD simulations (e.g., rigid-wall boundaries and simplified blood rheology), and further validation on datasets with more diverse physiological settings and clinical measurements such as 4D Flow MRI is needed. Third, although GHD provides a standardized mesh representation, translating the full pipeline to patient-specific reconstructions from medical imaging (CTA/MRA) may require additional preprocessing steps, which will be addressed in future work. 

\subsection{Clinical Implications}
From a clinical perspective, this work helps bridge the gap between CFD research and clinical decision-making. It can complement morphology-based risk scoring systems such as PHASES and ELAPSS by providing detailed hemodynamic parameters. Transient WSS and TAWSS show the location and extent of high- and low-shear regions, which are related to wall degeneration. RRT fields highlight zones associated with potential thrombosis. Although OSI remains numerically sensitive, the predicted OSI patterns capture the main oscillatory regions. In a practical workflow, the proposed surrogate could be applied to patient-specific meshes reconstructed from CTA, providing a comprehensive hemodynamic profile within seconds and could be integrated into routine diagnostic workflows to better inform treatment decisions for unruptured aneurysms.

\section{Conclusion}
In this work, we introduce a deep learning framework capable of predicting transient wall shear stress (WSS) fields across the full cardiac cycle directly from intracranial aneurysm (IA) surface meshes. By integrating geometric features with inlet waveform information through a graph-based architecture, our model provides real-time estimation of key biomechanical markers that traditionally require computationally expensive CFD simulations. Using the large-scale AneuG-Flow dataset, we demonstrate that the model captures the spatial distribution and temporal evolution of WSS, achieving sufficient agreement with CFD-derived ground truth.

A core contribution of our study is the use of inexpensive steady-state CFD simulations as data augmentation for transient prediction. Ablation experiments show that this strategy consistently improves performance—particularly when only limited pulsatile data are available—highlighting an effective path toward building high-fidelity surrogate models when transient datasets are inherently small and costly to generate. Our results further show that the predicted transient fields enable accurate computation of derived flow mechanics markers such as TAWSS and RRT, supporting their potential use in downstream biomechanical and clinical analyses.

Overall, this work demonstrates the feasibility of obtaining patient-specific, cardiac-cycle–resolved aneurysm hemodynamics in real time using a geometry-aware neural operator. Beyond improving computational efficiency, this approach may help bridge the long-standing gap between high-fidelity CFD research and routine clinical workflows by making biomechanical markers more accessible to physicians.

\section*{Acknowledgment}

This study was supported by the Imperial College startup funding and MOE-AcRF-Tier1-FRC-FY2024 Grant. We would also like to acknowledge that computational work involved in this study is partly supported by the National University of Singapore's IT Research Computing group under grant number NUSREC-HPC-00001.

\bibliographystyle{IEEEtran}
\bibliography{references}

\end{document}